\documentclass[letterpaper, 10 pt, conference]{ieeeconf} 
\IEEEoverridecommandlockouts                           

\title{\LARGE \bf
Fast Kinodynamic Bipedal Locomotion Planning with Moving Obstacles
}

\usepackage{cite}
\usepackage{hyperref}
\usepackage[pdftex]{graphicx}
\usepackage{epstopdf}
\usepackage{amsmath}
\usepackage{amssymb}
\usepackage{array}
\usepackage{wasysym}
\usepackage{bm}
\usepackage{units}
\usepackage[per-mode=symbol]{siunitx}
\usepackage{url}
\usepackage[usenames, dvipsnames]{color}
\usepackage{epstopdf}
\usepackage[ruled]{algorithm2e}
\usepackage{multirow}
\usepackage{booktabs}
\usepackage{lineno}
\usepackage{accents}

\newcommand*\myswitch{{\protect \includegraphics[width=0.8em]{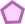}}}
\newcommand*\myapex{{\protect \includegraphics[width=0.8em]{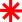}}}
\newcommand*\mycom{{\protect \includegraphics[width=0.8em]{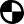}}}
\newcommand*\mybluecircle{{\protect \includegraphics[width=0.8em]{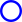}}}

\newcommand*\mybluebox{{\protect \includegraphics[width=0.8em]{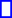}}}
\newcommand*\myredbox{{\protect \includegraphics[width=0.8em]{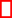}}}

\author{Junhyeok Ahn$^{1}$, Orion Campbell$^{1}$, Donghyun Kim$^{1}$, and Luis Sentis$^{2}$
\thanks{$^{1}$J.Ahn, O.Campbell, D. Kim are with the Graduate School of Mechanical Engineering, University of Texas at Austin, U.S.}
\thanks{$^{2}$Luis Sentis is associate professor in Aerospace Engineering and Engineering Mechanics, University of Texas at Austin, U.S.
{\tt\small lsentis@austin.utexas.edu}}
}

\begin{document}

\maketitle
\thispagestyle{empty}
\pagestyle{empty}

\begin{abstract}
In this paper, we present a sampling-based kinodynamic planning framework for a bipedal robot in complex environments. Unlike other footstep planning algorithms which typically plan footstep locations and the biped dynamics in separate steps, we handle both simultaneously. Three primary advantages of this approach are (1) the ability to differentiate alternate routes while selecting footstep locations \textit{based on the temporal duration of the route} as determined by the Linear Inverted Pendulum Model (LIPM) dynamics, (2) the ability to perform \textit{collision checking through time} so that collisions with moving obstacles are prevented without making a detour around the obstacles' entire trajectory, and (3) the ability to specify a minimum forward velocity for the biped. To generate a dynamically consistent description of the walking behavior, we exploit the Phase Space Planner (PSP) \cite{Kim:2017wv} \cite{Zhao:2012il}. To plan a collision-free route toward the goal, we adapt planning strategies from non-holonomic wheeled robots to gather a sequence of inputs for the PSP. This allows us to efficiently approximate dynamic and kinematic constraints on bipedal motion, to apply a sampling-based planning algorithm such as RRT or RRT*, and to use the Dubin's path \cite{Dubins:1957ho} as the steering method to connect two points in the configuration space. The results of the algorithm are sent to a Whole Body Controller \cite{Kim:2017wv} to generate a full body dynamic walking behavior. Our planning algorithm is tested in a 3D physics-based simulation of the humanoid robot Valkyrie. 
\end{abstract}
\section{INTRODUCTION}
\label{introduction}
We propose a new framework for fast kinodynamic locomotion planning for bipedal robots. Our planning algorithm is constructed based on a kinodynamic Rapidly-exploring Randomized Tree (RRT) \cite{Kuffner:2000gz} with a newly proposed method for approximating kinematic and dynamic constraints, which results in efficient computation, a complete solution and robust feed forward tasks for a Whole Body Controller (WBC) \cite{Kim:2017wv} to control a full body bipedal robot in a cluttered environment.

The foundation of our algorithm is an analytical solution to the Linear Inverted Pendulum Model (LIPM), which is a simplified dynamic model for bipedal robots. The LIPM not only provides a significantly reduced-order dynamically consistent manifold for planning but also generalizes bipedal locomotion so that it is agnostic to the specific robot configuration. In order to plan with LIPM dynamics, we utilize a Phase Space Planner (PSP) \cite{Zhao:2012il}. Different from previous walking pattern generators, PSPs uniquely require an inverted pendulum's desired forward velocity and the sagittal-plane foot placement of the next step. The PSP then generates the dynamically consistent locomotion parameters such as lateral-plane footstep location, the timing to switch to the following step and so on (see Appendix~\ref{sec:psp}) to ensure that the model proceeds along the desired heading. 
\begin{figure}
\centering
\includegraphics[width=0.9\columnwidth]{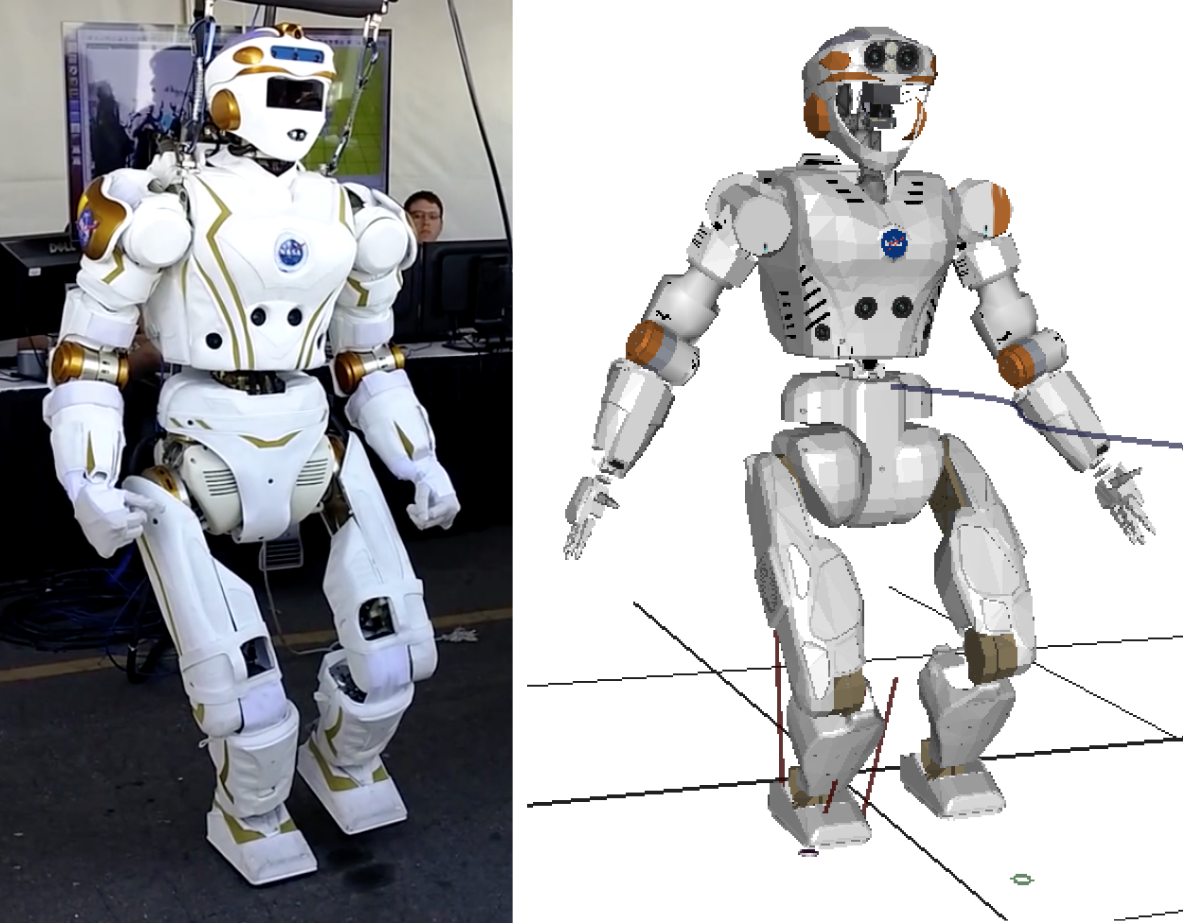}
\caption{\textbf {Type of humanoid platform our controller explores: } The left image shows NASA's Valkyrie humanoid robot, with 135.9 $\si[per-mode=symbol]{\kilo\gram}$ weight and 1.83 $\si[per-mode=symbol]{\meter}$ height. The right image shows our dynamic simulation of Valkyrie using the physics based simulator SrLib.}
\label{fig:valkyrie_figure}
\vspace{-1mm}
\end{figure} 

Since the PSP requires the sequence of the sagittal foot placements and forward velocities to generate a walking pattern, \cite{Zhao:2012il} manually specified these quantities and \cite{Kim:2017wv} exploited Reinforcement Learning to generate the required sequence. We adapt path planning strategies often used for non-holonomic wheeled robots to collect the required inputs for the PSP. At the same time, we use the wheeled robot's kinematic limitations as a rough approximation of the dynamic and kinematic constraints on bipedal motion. Additionally, this analogy transforms a high-dimensional discontinuous footstep selection problem into a low-dimensional continuous domain, which allows for the efficient use of RRT or RRT*. We compose the configuration space with the position and heading of the robot rather than using the full joint space like \cite{Harada:2009uh}, \cite{Hauser:2014il} or optimizing with the robot's full configuration like \cite{Posa:2014ez}, \cite{Mordatch:2012vn} which is typically highly computationally expensive. We introduce the Dubin's path \cite{Dubins:1957ho} as the steering method in our kinodynamic RRT which exactly connects any two states in an obstacle-free configuration space and contributes to the efficient performance of the RRT \cite{Kunz:2014bg}, \cite{Anonymous:2017ip}.

Lastly, we propose using a new metric for the kinodynamic RRT defined as the temporal duration of the movement, which can be calculated from the LIPM dynamics. Our formulation allows us to calculate the time that the robot will take to traverse any segment of the planned locomotion path. Generally, a metric in an RRT algorithm is used to determine the nearest neighbor on the tree and in the case of kinodynamic RRT, euclidean distance is inadequate as a metric because it does not account for the robot's dynamics. We believe that using the movement duration as the metric for our locomotion planner is effective because of the following capabilities. Choosing the nearest neighbor now corresponds to finding the neighbor from which you could start a movement and reach the sampled point the quickest. Also, this planner has capability to handle moving obstacles as long as the future obstacle locations are known (or can be well estimated) since collision checking will be evaluated for the future time calculated from the LIPM dynamics.

This capability is rare because the majority of prior work on bipedal locomotion planning divides the planning problem into two sequential phases(1) plan a finite sequence of footstep locations considering kinematic feasibility and obstacle avoidance, and (2) fit a dynamically consistent, continuous motion plan onto that footstep sequence. Because these two phases occur sequentially, there is no information about the dynamics of the robot in the first phase, and thus, no time-to-step information. By time-to-step, we mean the time it will take for the robot to reach a given location in the footstep sequence. For example, \cite{Deits:2015ch} generates an obstacle-free sequence of footstep locations using semi-definite programming while \cite{Perrin:2016dy} uses an RRT formulation. But only after the footstep sequence is obtained can they generate the robot's trajectory considering dynamics, whether that be for each joint individually or for the whole robot's Center of Mass (CoM). Thus, there is no way to know if a future collision with a moving object will occur because footstep locations are chosen (and collision checked) prior to knowing how long the movement will take.

In our study, we combine LIPM-PSP-kinodynamic RRT with a newly proposed process borrowed from wheeled robots, a steering method and a metric for locomotion planning that yield efficient and fast convergence in computation and can be applied to any kind of bipedal robot. Also, the planning algorithm provides completeness in the solution space and near optimality in terms of time through a rewiring process performed after the primary planning process. The results of this planning algorithm can be sent to a WBC \cite{Kim:2017wv} as tasks to generate whole body locomotion and are tested in a 3D physics-based simulation of the humanoid robot Valkyrie (Fig.~\ref{fig:valkyrie_figure}).

This paper is organized as follows. In Section~\ref{sec:problem_formulation}, we formulate kinodynamic RRT problem and propose new concepts. In Section~\ref{sec:planning_algorithm}, we describe how the PSP recursively propagates the LIPM dynamic integrated with kinodynamic planner. The results are presented in Section~\ref{sec:simulation_result}. Appendix~\ref{sec:psp} includes concept of the PSP and derivation of analytic solution of LIPM dynamics.

\section{Kinodynamic RRT Formulation}
\label{sec:problem_formulation}

To formulate our planning algorithm, we draw on the similarities between bipedal and wheeled robot navigation on a 2D plane because this provides several useful benefits. First, it converts a high-dimensional, discrete footstep planning problem into a low-dimensional, continuous domain with a well defined steering function that can be used for the RRT. Second, we can use the fairly simple limitation of a wheeled robot to conservatively approximate some of the complex kinematic and dynamic limits of the bipedal robot at the planning phase, especially the friction cone and balance limitations which prevent bipedal robots from instantaneously changing their direction of motion while walking. 

We construct the kinodynamic RRT with the configuration space composed of the position and heading of the robot and the non-holonomic constraint, adapting the wheeled robot description.
\begin{align}
\label{config}
\mathbf{q}&=[x,~y,~\theta]^T\\
\label{kinematic_x}
\dot{x}&=V\cos(\theta) \\
\label{kinematic_y}
\dot{y}&=V\sin(\theta) \\
\label{kinematic_u}
\dot{\theta}&=u\\
\label{bound_u}
\mid u \mid &\leq u_{max} \\
\label{col_free}
\mathbf{q}&\in C_{free}
\end{align}


Eq.~\eqref{config} shows the definition of the configuration space represented in global coordinates ($x^{\{g\}},~y^{\{g\}}$) and Eq.~\eqref{kinematic_x}--\eqref{kinematic_u} are the kinematic model of the wheeled robot. The solution trajectory should abide by the bounded input Eq.~\eqref{bound_u} and collision-free constraints Eq.~\eqref{col_free}. Note that a limitation of this formulation is that the robot is not allowed to turn in place, since we use a constant forward velocity, $V$. If desired, such movements must be planned separately. Two of these wheeled robot parameters are used to approximate the bipedal robot's reachability and dynamics. Specifically, the constrained input (Eq.~\eqref{bound_u}) enforces a  minimum turning radius for the biped, $r_{min}$, and $V$ sets a lower bound for the forward CoM velocity.

The classic RRT formulation plans a collision free path through the configuration space by building a tree of possible movements starting from the initial configuration until a branch of the tree can be connected to the goal configuration. A tree, $\mathbb{Q}_{tree}$, is comprised of a set of nodes with connections that link the nodes into branches, or sequences. The structure of a tree is such that a node, $\mathbf{q}_i$, can have many children, but only one parent, $\mathbf{q}_{p(i)}$. Sequences of nodes, or branches, represent potential navigational paths through the configuration space. 

We adapted a variant of RRT called RRT-Connect \cite{Kuffner:2000gz} for use with the wheeled robot described above. We summarize the procedure from a high level as follows: (0) initialize the tree with a single node at the starting configuration, (1) sample a random point $\mathbf{q}_s$ in the configuration space, (2) connect each existing node on the tree to the newly sampled point using a constrained steering function, but ignoring obstacles, (3) choose the closest node $\mathbf{q}_{nn}$ to the sampled point (commonly called the "nearest neighbor") by measuring and comparing each of these path lengths according to some metric, (4) extend the tree by placing new nodes ($\mathbf{q}_{nn^{+1}},\mathbf{q}_{nn^{+2} },~...,~\mathbf{q}_s$) along the path connecting the "nearest neighbor" on the tree to the sampled point until a collision occurs or the sampled point is reached, (5) repeat steps (1)--(4), until the goal configuration has been connected to the starting configuration with a valid path. With some frequency, one should substitute the goal configuration for the randomly sampled point in step (1) so that occasional attempts are made to complete the tree. Once the goal configuration has been successfully connected to the tree, the solution is the sequence of nodes which connect the starting configuration to the goal configuration.
\begin{figure}[htpb]
    \centering
    \includegraphics[width=0.8\columnwidth]{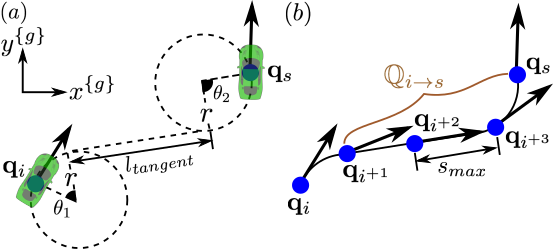}
    \caption{\textbf{ Dubin's path as the steering function and intermediate nodes} : (a) illustrates two nodes connected through the shortest path among the six families of Dubin's path, represented as dotted line. (b) shows extended tree by placing new intermediate nodes on the Dubin's path.}
    \label{fig:steering}
\end{figure}

In our planner, the concept of a Dubin's path \cite{Dubins:1957ho} is introduced as the steering function to efficiently connect two points in the configuration space (e.g. $\mathbf{q}_i,~\mathbf{q}_s$ in Fig.~\ref{fig:steering}(a)) with the shortest navigational path while following the kinematic constraint of the wheeled robot (Eq.~\eqref{bound_u}) and ignoring obstacles. Among the six families of Dubin's car solution paths comprised of straight lines and circular arcs with radius $r_{min}$, we compare the total length, 
\begin{equation}
l_{i \rightarrow s} = r_{min}(\mid\theta_{1}\mid+\mid\theta_{2}\mid)+l_{\text{tangent}},
\end{equation}
in Fig.~\ref{fig:steering}(a), of each of the six candidates and choose the shortest path that connects the two points. Once the minimum $l_{i \rightarrow s}$ is computed for every node in $\mathbb{Q}_{tree}$, we define $\mathbb{Q}_{\text{closest}}$ as the set of the $k=20$ closest nodes. For the path connecting each $\mathbf{q}_i \in \mathbb{Q}_{\text{closest}}$ to $\mathbf{q}_s$, we locate the minimum number of kinematically reachable intermediate nodes $\mathbb{Q}_{i \rightarrow s} \triangleq (\mathbf{q}_{i^{+1}},~\mathbf{q}_{i^{+2}},~...,~\mathbf{q}_s)$ spaced evenly along the path (Fig.~\ref{fig:steering}(b)) so that the path length between any two sequential intermediate nodes (e.g. $\mathbf{q}_{i^{+1}}$ and $\mathbf{q}_{i^{+2}}$) is less than or equal to $s_{max}$ where $s_{max}$ is a conservative upper bound for the biped's step length. Thus, for each $\mathbf{q}_i \in \mathbb{Q}_{\text{closest}}$, we have identified a potential new branch $\mathbb{Q}_{i \rightarrow s}$ that could be added to $\mathbb{Q}_{tree}$, and we will develop a metric which we will use to select the nearest neighbor node $\mathbf{q}_{nn}$ and its branch $\mathbb{Q}_{nn\rightarrow s} \triangleq (\mathbf{q}_{nn^{+1}},~\mathbf{q}_{nn^{+2}},~\cdots,~\mathbf{q}_{s})$ among the 20 branches to append to the tree.

However, a sequence of nodes that trace a path from the initial configuration to the goal configuration does not contain enough information for a bipedal robot to actually walk through these waypoints. Thus, we introduce the PSP (Algorithm~\ref{code:PSP}) which operates on the LIPM to solve for locomotion parameters like the dynamically consistent step location, stance-foot switching time, and the LIPM state after taking the specified step that connects the two nodes. With some information about the initial LIPM state and a sequence of nodes, the PSP can recursively operate on each pair of nodes in the sequence to propagate the LIPM along the path, and generate the information needed to produce dynamically consistent walking motion with a WBC. Additionally, since the PSP calculates the elapsed time between nodes based on the LIPM's dynamics, we can use this information both (1) to judge alternate routes according to which path is the \textit{fastest} rather than simply using the the shortest path, and (2) to perform \textit{collision checking through time}. The first of these capabilities will be used as our metric to select one of the $k=20$ potential new branches, $\mathbb{Q}_{i \rightarrow s} ~\forall~ \mathbf{q}_i \in \mathbb{Q}_{\text{closest}}$, to be appended to $\mathbb{Q}_{tree}$. One assumption here is that the fastest path will always appear as one of the 20 shortest length paths, but we have found that this is basically always the case. The second of these capabilities means that if there are moving objects in the environment whose motion is defined or can be estimated, then this planner can accurately detect collisions in the future. Thus, unlike other planners, which may have to avoid the entire path of moving obstacles, since they select footstep locations prior to solving the biped's dynamics, this planner plans paths that cross that of moving obstacles, as long as they are both not in the same place at the same time.

Based on all of notation and concepts introduced so far, we can summarize the locomotion planning problem in a cluttered environment as follows: the objective is to find a solution sequence $\mathbb{Q}_{sol} \triangleq \mathbb{Q}_{\text{start} \rightarrow \text{goal}}$ connecting $\mathbf{q}_{start}$ and $\mathbf{q}_{goal}$ and the corresponding locomotion parameters which follow the waypoints defined by that sequence, given the parameters $r_{min},~s_{max},\text{ and}~V$. We limit the exploration of the space to be within a given configuration boundary $(\mathbf{q}_{min},~\mathbf{q}_{max})$ and we require the solution to avoid any collisions with obstacles in the space (defined by $C_{free}(t)$). 

\section{Propagating LIPM Dynamics}
\label{sec:planning_algorithm}
\begin{figure}[b]
    \centering
    \includegraphics[width=0.9\columnwidth]{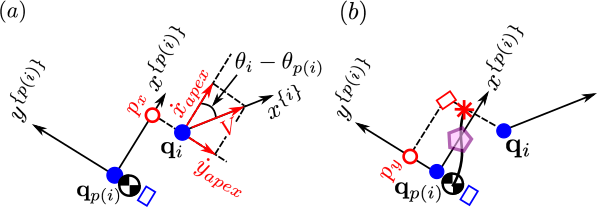}
    \caption{\textbf{ LIPM dynamics propagation }(a) illustrates calculating an input vector for PSP. (b) shows the output vector of PSP. These input and output vectors compose of the locomotion parameter $\mathbf{m}_{i}$ and describing the LIPM dynamics for the step at node $\mathbf{q}_{i}$.}
    \label{fig:in_and_out}
\end{figure}
In this section, we describe the method for recursively propagating the LIPM dynamics using the PSP along a sequence of nodes $\mathbb{Q}_{i \rightarrow s}$. This will allow us to algorithmically evaluate the dynamical consequences of each of the 20 potential new branches which were isolated in the previous section. From each of these new branches, we would like to determine the total duration of a biped's movement along that branch, or $t_{i \rightarrow s}$. The branch with the smallest total duration will be selected for addition to $\mathbb{Q}_{tree}$.

First, we describe how we will track the dynamics of the biped throughout the tree. The locomotion parameters including the LIPM state for a step that connects adjacent nodes $\mathbf{q}_{p(i)}$ and $\mathbf{q}_i$ are collectively stored in a variable $\mathbf{m}_i$. Since each node has a unique parent, and we calculate an $\mathbf{m}_i$ for every $\mathbf{q}_i$, we can think of the set of all $\mathbf{m}_i$'s, or $\mathbb{M}_{tree}$, as forming a "mirror" tree which captures the bipedal walking dynamics along the navigational waypoints in $\mathbb{Q}_{tree}$. The $\mathbf{m}_i$ vector can be split into the PSP input and output, and is defined as follows,
\begin{equation}
\label{eq:in_and_out}
\begin{split}
\mathbf{m}_i& \triangleq [\mathbf{m}_i^{in},~\mathbf{m}_i^{out}]^T\\
\mathbf{m}_i^{in}& \triangleq [p_{x},~\dot{x}_{apex},~\dot{y}_{apex}]^T\\
\mathbf{m}_i^{out}& \triangleq [t_{switch},~t_{apex},~p_{y},~y_{apex}]^T
\end{split}
\end{equation}
where $[p_x,~p_y]^T$ is the dynamically consistent and reachable step location, defined in the local frame located and oriented with $\mathbf{q}_{p(i)}$, shown as \myredbox ~in Fig.~\ref{fig:in_and_out}(b); $[x_{apex},~y_{apex}]^T$ and $[\dot{x}_{apex},~\dot{y}_{apex}]^T$ are the CoM position and velocity, again in the local frame, at the apex (\myapex) of the step; $t_{switch}$ is the time elapsed from the previous apex (\mycom) to the stance-foot switching (\myswitch), and $t_{apex}$ is the time elapsed between the stance-foot switching (\myswitch) and the next apex (\myapex). Note that the apex of a step occurs when the CoM is positioned directly above the stance foot in the sagittal plane so that $x_{apex} = p_x$. The two durations $t_{switch}$ and $t_{apex}$ sum to yield the total duration of the step, and are illustrated in Fig. \ref{fig:PSP}. Thus, once all $\mathbf{m}_i$'s are known for any consecutive sequence of steps, $\mathbb{M}_{i \rightarrow s}$, the total duration of the sequence can be obtained merely by summing $t_{switch}$ and $t_{apex}$ from each of the $\mathbf{m}_i$'s in the sequence:
\begin{equation}
\label{t_dur_eqn}
t_{i \rightarrow s} = \sum_{n=i^{+1}}^{s}\mathbf{m}_n.t_{switch} + \mathbf{m}_n.t_{apex}
\end{equation}
Note that for generality, we use notation $\mathbf{q}_{i},~\mathbf{q}_{p(i)}$ in Fig.~\ref{fig:in_and_out} and Eq.~\eqref{eq:in_and_out} but the $\mathbb{Q}_{i \rightarrow s}$ and $\mathbb{M}_{i \rightarrow s}$ start with $\mathbf{q}_{i^{+1}}$ and $\mathbf{m}_{i^{+1}}$.

With this notation introduced, we can move on to describe the procedure for propagating the locomotion parameters and LIPM dynamics through a "mirror" branch $\mathbb{M}_{i \rightarrow s}$ of the configuration space branch $\mathbb{Q}_{i \rightarrow s}$. In this situation, $\mathbb{Q}_{i \rightarrow s}$ has been populated with a sequence of configurations, and we will start at the beginning of the branch, where the parent node ($\mathbf{q}_{i}$) of the new node ($\mathbf{q}_{i^{+1}}$) is an existing node on the tree. As a result, $\mathbf{m}_{p(i^{+1})}$ (or $\mathbf{m}_{i}$) is known. However, to persist generality, we will denote the new node whose locomotion parameters are to be computed as $\mathbf{q}_{i}$ and its parent node whose locomotion parameters are already known as $\mathbf{q}_{p(i)}$. We will continue from this point by focusing on a pair of nodes $\mathbf{q}_{p(i)}$ and $\mathbf{q}_i$ with the objective being to populate $\mathbf{m}_i$, noting that the same procedure for the first pair is to be repeated for each pair in the sequence until one reaches the last pair in the branch, ending with $\mathbf{q}_s$. We calculate $\mathbf{m}_i^{in}$ as follows
%
\begin{equation}
\begin{split}
\label{collect_input}
\begin{bmatrix} \mathbf{m}_i.p_{x} \\ \_\_ \\ \_\_ \end{bmatrix} &= T^{g}_{p(i)}\begin{bmatrix} \mathbf{q}_{i}.x \\ \mathbf{q}_{i}.y \\ 1 \end{bmatrix} \\
\mathbf{m}_i.\dot{x}_{apex} &= V\cos(\mathbf{q}_i.\theta-\mathbf{q}_{p(i)}.\theta) \\
\mathbf{m}_i.\dot{y}_{apex} &= V\sin(\mathbf{q}_i.\theta-\mathbf{q}_{p(i)}.\theta)
\end{split}
\end{equation}
where \_\_ indicates that the value is not used, $T^{g}_{p(i)}$ is an SE(2) transformation matrix from the global coordinate frame $\{g\}$ to the $\{p(i)\}$ coordinate frame. Once $\mathbf{m}^{in}_{i}$ is calculated, the PSP  computes $\mathbf{m}^{out}_{i}$ with the pair of the locomotion parameters $\mathbf{m}_{p(i)}$ and $\mathbf{m}^{in}_{i}$. The PSP algorithm is described in more detail in Appendix~\ref{sec:psp}. Notice that the locomotion parameters $\mathbf{m}_{i}$ calculated used in PSP are represented with respect to the parent node's local frame $\{p(i)\}$. One should transform $\mathbf{m}_{i}$ including foot placement and LIPM state into its local frame $\{i\}$ and compute $\mathbf{m}^{\{i\}}_{i}$ in order to compute locomotion parameters for the next node. Since $\mathbf{m}_{i}$ is augmented position and velocity vector in Cartesian space, it can be transformed by the augmented transformation matrix.
\begin{algorithm}[htbp]
    \SetKwFunction{Collect}{Compute\_PSP\_Input}
    \SetKwFunction{Append}{Append}
    \SetKwFunction{Trans}{Transform}
    \SetKwFunction{Time}{Compute\_Duration}
    \SetKwFunction{PSP}{PSP} 
	
	\KwIn{$\mathbb{Q}_{i\rightarrow s}$}
	\KwResult{$\mathbb{M}_{i\rightarrow s},~t_{i\rightarrow s}$}
	\vspace{1.5mm}
    \For{\rm{each} $\mathbf{q}_{i} \in \mathbb{Q}_{i \rightarrow s}$ }{
    	$\mathbf{m}^{in}_{i} \gets$ \Collect{$\mathbf{q}_{p(i)},~\mathbf{q}_{i}$} \\ \tcp*[f]{Eq.~\eqref{collect_input}} \\
        $\mathbf{m}^{\{p(i)\}}_{p(i)} \gets$ \Trans{$\mathbf{m}_{p(i)}$} \vspace{1.5mm} \\
    	$\mathbf{m}^{out}_{i} \gets$ \PSP{$\mathbf{m}^{\{p(i)\}}_{p(i)},~\mathbf{m}^{in}_{i}$} \tcp*[f]{Algorithm~\ref{code:PSP}} \vspace{1.5mm} \\
        $\mathbb{M}_{i\rightarrow s} \gets$ \Append{$\mathbf{m}^{in}_{i},~\mathbf{m}^{out}_{i}$} \\
    }
	$t_{i\rightarrow s} \gets$ \Time{$\mathbb{M}_{i \rightarrow s}$} \tcp*[f]{Eq.~\eqref{t_dur_eqn}} \\
 \caption{Computation of $~\mathbb{M}_{i \rightarrow s}$ and $t_{i \rightarrow s}$ for a given $\mathbb{Q}_{i \rightarrow s}$}
 \label{code:seg}
\end{algorithm}

At this point, one can use the above recursive procedure along with Algorithm~\ref{code:seg} to determine $\mathbb{M}_{i \rightarrow s}$ and $t_{i \rightarrow s}$ for each of the $k=20$ potential new branches, $\mathbb{Q}_{i \rightarrow s} ~\forall~ \mathbf{q}_i \in \mathbb{Q}_{\text{closest}}$. As previously mentioned, the fastest route is selected and can now be identified as the sequence connecting the "nearest neighbor" on the existing tree to the sampled node, or $\mathbb{Q}_{nn \rightarrow s}$. Since the time of arrival for each node in this sequence is known, this branch is ready to be collision checked through time to determine how much of the branch will be appended to $\mathbb{Q}_{tree}$. The collision checking process will be explained in further detail in the illustrated example that follows. 
\begin{figure*}[htbp]
\centering
\includegraphics[width=1.9\columnwidth]{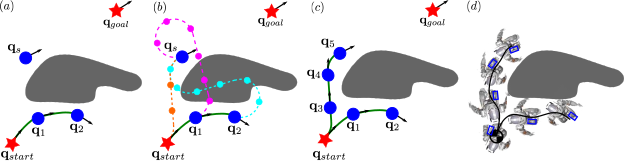}
\caption{\textbf{Kinodynamic locomotion planning: }(a) shows a starting noded $\mathbf{q}_{start}$, a goal node $\mathbf{q}_{goal}$ and a randomly sampled node $\mathbf{q}_{s}$. (b) shows the shortest Dubin's path from each $\mathbf{q}_{i}$ $\in$ $\mathbb{Q}_{sol}$ to $\mathbf{q}_{s}$. The steering function collects $\mathbb{Q}_{i \rightarrow s}$ and computes the sequence of locomotion parameters $\mathbb{M}_{i \rightarrow s}$ on each Dubin's path through Algorithm~\ref{code:seg}. (c) illustrates $\mathbb{Q}_{start \rightarrow s}$ is chosen as the nearest neighbor based on time metric and appended. After checking collision through time, $\mathbb{Q}_{start \rightarrow s}$ and $\mathbb{M}_{start \rightarrow s}$ are appended to $\mathbb{Q}_{tree}$ and $\mathbb{M}_{tree}$. (d) shows the actual robot's CoM trajectory to be operated (black solid line) and the sequence of foot placements.}
\label{fig:one_loop} 
\end{figure*}

The overall planning algorithm is described in Algorithm~\ref{code:planning_algo}, Fig.~\ref{fig:one_loop} and we demonstrate one cycle of while loop for clarification. In Fig.~\ref{fig:one_loop}(a), $\mathbf{q}_{start}$ and $\mathbf{q}_{goal}$ are illustrated. The green trajectory passing through $\mathbf{q}_{1},~\mathbf{q}_{2}$ and corresponding $\mathbf{m}_{1},~\mathbf{m}_{2}$ illustrates the existing node sequences which have been added to the two trees ($\mathbb{Q}_{tree}$ and $\mathbb{M}_{tree}$) in previous iterations. $\mathbf{q}_{s}$ is a randomly sampled node within the boundaries of the configuration space. 

In Fig.~\ref{fig:one_loop}(b), the steering function chooses the shortest Dubin's path (the dotted lines) from every $\mathbf{q}_{i} \in \mathbb{Q}_{tree}$ to $\mathbf{q}_{s}$, generates $\mathbb{Q}_{i \rightarrow s}$ with intermediate nodes for 20 closest nodes among Dubin's path, and iteratively computes the sequence of locomotion parameters $\mathbb{M}_{i \rightarrow s}$ and $t_{i \rightarrow s}$. This can be done by executing Algorithm~\ref{code:seg} repeatedly. Then we choose the path connecting the nearest neighbor to the sampled node by comparing each of the $t_{i \rightarrow s}$'s and proceed to collision checking. We stipulate that a particular node $\mathbf{q}_i$ and $\mathbf{m}_i$ is collision-free if there is no intersection between any obstacles' bounding boxes at the time of arrival $t_{i}$ and a circle centered at a foot placement $(\mathbf{m}_{i}.p_{x},~\mathbf{m}_{i}.p_{y})$ with a radius defining a circular safety margin around the robot. $t_i$ can be computed by adding up time information in locomotion parameters from $\mathbf{q}_{start}$ to $\mathbf{q}_{i}$. If a node is determined to collide with an obstacle or the configuration space boundary, then that node and all future nodes in that sequence are pruned. Finally, the pruned sequences $\mathbb{Q}_{i \rightarrow s}$ and $\mathbb{M}_{i \rightarrow s}$ are appended to $\mathbb{Q}_{tree}$ and $\mathbb{M}_{tree}$ as shown in Fig.~\ref{fig:one_loop}(c). Fig.~\ref{fig:one_loop}(d) shows the actual robot executing locomotion based on a sequence of locomotion parameters. The black solid line represents CoM trajectory and \mybluebox ~shows the foot placements.
\begin{algorithm}[htbp]
	\SetKwFunction{RandomSample}{Random\_Sample}
    \SetKwFunction{Steering}{Steering}
	\SetKwFunction{Seg}{Compute\_Segment}
    \SetKwFunction{NN}{Get\_Nearest\_Neighbor}
    \SetKwFunction{RW}{Rewiring}
    \SetKwFunction{Prune}{Prune\_For\_Collision}
    \SetKwFunction{PruneSol}{Get\_Solution}
    \SetKwFunction{append}{Append}
	\KwIn{$\mathbf{q}_{start}$, $\mathbf{q}_{goal}$, $\mathbf{q}_{min}$, $\mathbf{q}_{max}$, $C_{free}(t)$, $r_{min}$, $s_{max}$, $V$}
	\KwResult{$\mathbb{Q}_{sol},~\mathbb{M}_{sol}$}
	\vspace{1.5mm}
	\While{$\mathbf{q}_{goal} \not \in \mathbb{Q}_{tree}$}
    {
		$\mathbf{q}_{s} \gets$ \RandomSample{$\mathbf{q}_{min},~\mathbf{q}_{max}$} \\ \tcp*[f]{Fig.~\ref{fig:one_loop}(a)} \\
        $\mathbb{Q}_{\text{closest}} \gets$ \Steering{$\mathbb{Q}_{tree},\mathbf{q}_{s},r_{min},s_{max},V$}\\ \tcp*[f]{Fig.~\ref{fig:one_loop}(b)} \\
		\For{ \rm{each} $\mathbf{q}_{i} (\in \mathbb{Q}_{\text{closest}})$}{
        	$\mathbb{M}_{i \rightarrow s},~t_{i \rightarrow s} \gets$ \Seg{$\mathbb{Q}_{i \rightarrow s}$} \\ \tcp*[f]{Algorithm~\ref{code:seg}} \\
        }
        $\mathbb{Q}_{nn \rightarrow s},~\mathbb{M}_{nn \rightarrow s} \gets$ \NN \\ \tcp*[f]{Fig.~\ref{fig:one_loop}(c)} \\
        $\mathbb{Q}_{tree},~\mathbb{M}_{tree} \gets$ \append{$\mathbb{Q}_{nn \rightarrow s},~\mathbb{M}_{nn \rightarrow s}$} \\
	}
    $\mathbb{Q}_{sol},~\mathbb{M}_{sol} \gets$ \PruneSol{$\mathbb{Q}_{tree},~\mathbb{M}_{tree}$} \\
    $\mathbb{Q}_{sol},~\mathbb{M}_{sol} \gets$ \RW{$\mathbb{Q}_{sol},~\mathbb{M}_{sol}$} \\
 \caption{Kinodynamic locomotion planning}
 \label{code:planning_algo}
\end{algorithm}

Once the planner successfully connects $\mathbb{Q}_{tree}$ to $\mathbf{q}_{goal}$ and compute mirroring $\mathbb{M}_{tree}$, the unused branches are thrown away, leaving only a single sequence of steps, defined by $\mathbb{Q}_{sol}$ and $\mathbb{M}_{sol}$. Although this motion could be used, it tends to wander slightly through the space, as is typical of RRT solutions. Thus, we use a rewiring process to smooth and optimize the planned motion. To do this, we randomly select two different nodes $\mathbf{q}_{m},~\mathbf{q}_{n} ~(\in \mathbb{Q}_{sol},~m \neq n)$ and apply Algorithm~\ref{code:seg} which attempts to connect the two nodes directly with a new Dubins path solution and the corresponding sequence of new nodes, $\mathbb{Q}_{m \rightarrow n}^{alt}$ and $\mathbb{M}_{m \rightarrow n}^{alt}$. If the new, alternate nodes are collision free and $t_{m \rightarrow n}^{alt} \leq t_{m \rightarrow n}^{orig}$ then the original nodes between $\mathbf{q}_m$ and $\mathbf{q}_n$ are replaced with the new, alternate nodes $\mathbb{Q}_{m \rightarrow n}^{alt}$ and $~\mathbb{M}_{m \rightarrow n}^{alt}$. Since the locomotion parameters and LIPM state at $\mathbf{m}_{n}$ has changed as a result of this replacement, all future nodes downstream of $\mathbf{m}_n$ must be recalculated via Algorithm~\ref{code:PSP}. So, $\mathbb{M}_{n \rightarrow goal}$ are also replaced based on the update to the locomotion parameters at node $n$. Repeating this rewiring process successively improves the solution until it closely hugs corners and dodges obstacles.

\section{Simulation result}
\label{sec:simulation_result}
\begin{figure*}[htbp]
\centering
\includegraphics[width=1.9\columnwidth]{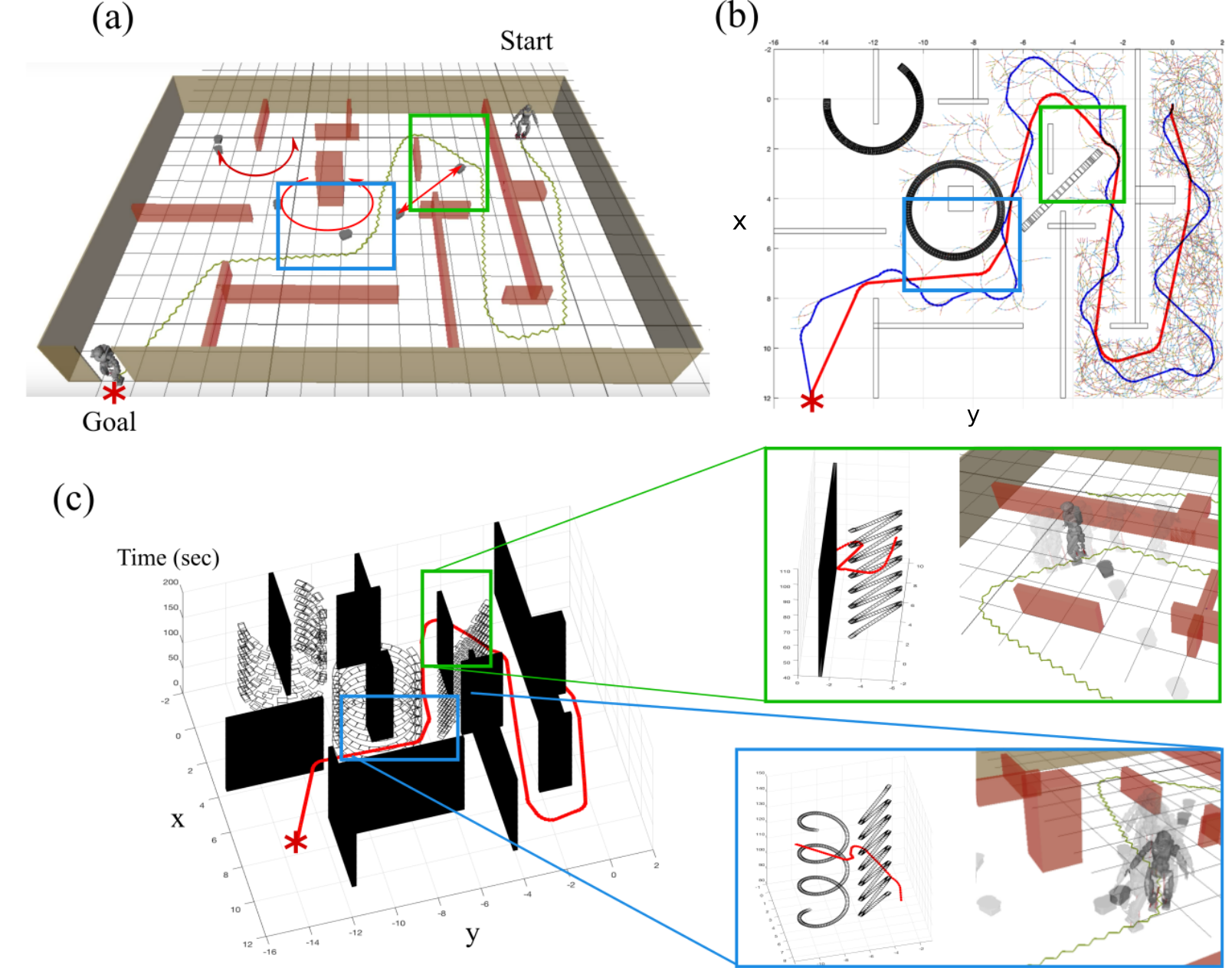}
\caption{\textbf{Environment Setup and Planned Path.} (a) shows the robot's initial state and goal position (\myapex). The room separated by red walls and three mobile robots are moving around with a regular speed. Two are revolving around a certain point and the other is moving linearly. (b) shows the top view of (c). After the planner finds the solution path (blue), the rewiring process smooths out the path (red). (c) shows the graph in time domain as well as Cartesian space. We could see the static walls stand still over the time and dynamic obstacles move through time. Over three figures, the locations indicated by green and blue squares are identical.} \label{fig:result_tot} 
\end{figure*}
To validate the proposed algorithm, we test it with a full
human-sized bipedal robot, Valkyrie, in a dynamic simulator, srLib\footnote{Seoul National University Robotics Library. Open-source \url{http://robotics.snu.ac.kr/srlib/}}. In the simulation, Valkyrie's starting location is in the right upper corner of a maze in an $18 \times 14 \si{\meter}$ room. The maze (shown in Fig.~\ref{fig:result_tot}(a)) is formed with red walls and has three mobile robots (shown as small gray boxes) that move through known trajectories (shown as red arrows). The mission is to walk to the door at the left bottom corner of the room while avoiding static and moving obstacles. We show the problem specification in Table~\ref{tab:problem}.

\begin{table}[htbp]
\centering
\begin{tabular}{ |c || l | }
\hline
Sampling&$\mathbf{q}_{min}=[-2,~12,0]^T$\\
Boundary&$\mathbf{q}_{max}=[-16,~2,2\pi]^T$\\ 
\hline
~& $\mathbf{m}_{start}=[p_x=0.195,~p_y=-0.13,$\\
~& \qquad \qquad $~{x}_{apex}=0.195,~\dot{x}_{apex}=0.04,$\\
~& \qquad \qquad $~{y}_{apex}=-0.052,~\dot{y}_{apex}=0,$\\
Mission& \qquad \qquad $~t_{switch}=0,~t_{apex}=0]^T~~$\\
~& ($\mathbf{m}_{start}$ has no parent, so expressed in the global frame)\\
~& $\mathbf{q}_{start}=[0.195,~-0.052,~0]^T$\\
~& $\mathbf{q}_{goal}=[12,~-14.5,~0]^T$\\
 \hline
 ~& $s_{max}=0.17$\\
 Parameters&$r_{max} = 0.5 $\\
 &$V=0.3$\\
 \hline
 Obstacles&Static obs. (red boxes)\\
 Movement&Dynamic obs. (red arrows)\\
  \hline
\end{tabular}
\vspace{1mm}
\caption{Problem Formulation of Fig.~\ref{fig:result_tot}}
\label{tab:problem}
\vspace{-5mm}
\end{table}


The proposed planner successfully finds the solution route to the goal location in 70 $\si{\second}$ at most including rewiring process for 108 steps. In Fig.~\ref{fig:result_tot}(b)), the blue trajectory illustrates the original $\mathbb{Q}_{sol}$ and the red trajectory shows the rewired solution. The multicolored dots in the figure are the nodes on $\mathbb{Q}_{tree}$ found during the exploration process which were not part of the solution sequence. Since the algorithm is based on random sampling, the computation time and final solution from each trial is not identical. As shown in Fig.~\ref{fig:result_tot}(c), the solution can be visualized as navigating through the time domain as well as Cartesian space. Note that static obstacles such as the walls remain stationary as time increases, while the moving obstacles do not. When the biped passes the moving obstacle revolving around the center of the maze (see the blue box magnification), some nodes are included in $\mathbb{Q}_{tree}$ which cross paths with the mobile robot while it is on the other side of the circle. To control the full body motion of Valkyrie, we generate a CoM task and foot task for the WBC proposed in \cite{Kim:2017wv} from the solution sequence. Note this planning situation is significantly more complicated than most practical planning problems, which usually plan footstep sequences over much smaller distances. 

Despite the fact that the above situation is somewhat unrealistically complex for practical bipedal planning problems, we chose to test the algorithm with this situation because it demonstrates some of the strengths and capabilities of our algorithm. In the process of solving that problem, the planner typically had on the order of $15000$ nodes in $\mathbb{Q}_{tree}$ when it found a solution. However, this algorithm works as well for simpler planning problems as it does for the highly complex one presented above. When there are fewer tight clearance passages between the starting location and the goal location as is the case in most practical biped navigation planning problems, the planner often can find and rewire a solution with $100$'s of steps in fractions of a second.

\section{Conclusion and Discussion} 
In this paper, we present a novel locomotion planning framework for biped robots with the combination of LIPM-PSP-kinodynamic RRT. We adapt a non-holonomic wheeled robot description to approximate some of the kinodynamic limits of bipeds and exploit a sampling based approach to exploring the configuration space. Within our kinodynamic RRT, we exploit the Dubin's path as a steering method and propose a new elapsed time metric to select between alternate routes. The proposed planning algorithm accounts for kinematic reachability, moving obstacle avoidance, and dynamic consistency. The output of our planner can be directly used as WBC tasks that generate robust navigation and locomotion behavior.

The planning algorithm in this paper only addresses forward walking behaviors. However, one possible avenue of extending this work could be to include other bipedal capabilities to generate various combinations of walking behaviors including side steps, turning in place, or walking backward.

As another possible extension, we may modify this framework so that it can be used for real-time replanning in a complex environment where the movement of obstacles is \textit{not} known a priori. Based on current observations of mobile obstacles, the robot could generate a probabilistic map of the future location of obstacles. Then, by choosing a probabilistic intolerance for collisions, one could replan routes in real-time to avoid collisions with moving objects or humans.

\appendices
\section{Phase Space Planner}
\label{sec:psp}
\begin{figure}[htpb]
    \centering
    \includegraphics[width=0.9\linewidth]{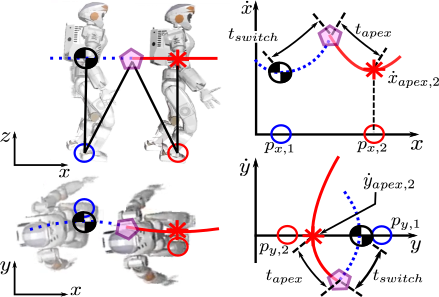}
    \caption{\textbf{Consecutive foot steps and CoM trajectories in Cartesian and phase space} : \mycom ~and \mybluecircle ~are a current CoM and foot step, followed by red color representations. For each stances, a CoM \myapex ~above sagittal foot placement is called apex and the intersection \myswitch ~is represented by switching. $t_{switch}$ and $t_{apex}$ are defined as time duration from \mycom ~to \myswitch ~and from \myswitch ~to \myapex.}
    \label{fig:PSP}
\end{figure}

PSP generates effective step switching information using simplified models such as the LIPM. In Fig.~\ref{fig:PSP}, we show phase plots across multiple walking steps of the CoM sagittal and lateral phase portraits based on LIPM dynamics. In the sagittal plane, the path consists of connected parabolas, while in the lateral plane, the walking path follows semi-periodic parabolas in a closed cycle. For convenience, we will use $x$ for the sagittal plane and $y$ for the lateral plane.

Compared to other locomotion algorithms such as \cite{Englsberger:20k5jp}, \cite{Kajita:2003uh}, PSP computes information regarding to footstep changing (e.g. $t_{switch},~t_{apex},~p_{y,2}$ and $y_{apex,2}$) with a given forward step location $p_{x,2}$ and an apex velocity $\dot{x}_{apex,2}$ and $\dot{y}_{apex,2}$. In the $x$ phase plot, we can see the given current CoM state \mycom ~and the apex state \myapex ~uniquely define switching state \myswitch,~ $t_{switch}$ and $t_{apex}$. These two timing values are used to find the next lateral step location $p_{y,2}$ and lateral CoM position $y_{apex,2}$ at apex. In summary, considering a one step ahead plan with a current CoM state and foot stance (\mycom ~ and \mybluecircle) and desired future states $[p_{x,2},~\dot{x}_{apex,2},~\dot{y}_{apex,2}]^T$, PSP finds locomotion parameter vector $[t_{switch},~t_{apex},~p_{y,2},~y_{apex,2}]^T$ to generate walking pattern for the following step.

When we constrain LIPM dynamics to a piecewise linear height surface, $z = a(x-p_x) + b$, we can find $t_{switch}$ and $p_y$ without numerical integration and bisection search because the system of equations becomes linear, resulting in the following CoM behavior:
\begin{equation}
\label{eq:x_state}
\begin{split}
x(t) & = A e^{\omega t} + B e^{-\omega t} + p_x, \\
\dot{x}(t) & = \omega (A e^{\omega t} - B e^{-\omega t} ),
\end{split}
\end{equation}
where,
\begin{equation}
\begin{split}
\omega &= \sqrt{\frac{g}{a p_x + b}}, \\
A &= \frac{1}{2}\Big( (x_{0} - p_x) + \frac{1}{\omega}\dot{x}_{0}   \Big), \\
B &= \frac{1}{2}\Big( (x_{0} - p_x) - \frac{1}{\omega}\dot{x}_{0}   \Big).
\end{split}
\end{equation}
Note that this equation is the same for the $y$ direction. Based on Eq.~\eqref{eq:x_state}, we can find an analytical solution for PSP, summarized in Algorithm \ref{code:PSP}. $~\mathbf{x}_{1}$, $\mathbf{y}_{1}$, $\mathbf{x}_{apex,2}$, and $\mathbf{x}_{switch}$ are vector quantities corresponding to the variables $(x_1,\dot{x}_1)$, $(y_1,\dot{y}_1)$, $(x_{apex,2}, \dot{x}_{apex,2})$, and $(x_{switch},\dot{x}_{switch})$.
\begin{algorithm}
\caption{Computation of locomotion parameters}\label{code:PSP}
\SetKwFunction{FindXSwitchingState}{Get\_Switching}
\SetKwFunction{GetTimeAtState}{Get\_Time}
\SetKwFunction{FindPy}{Find\_Py}
\SetKwFunction{Integration}{GetState}
\KwIn{ $[\mathbf{x}_1,~\mathbf{y}_1,~p_{x,1},~p_{y,1},~p_{x,2}~\dot{x}_{apex,2},~\dot{y}_{apex,2}]^T$}
\KwResult{ $[t_{switch},~t_{apex},~p_{y,2},~y_{apex,2}]^T$ } \vspace{1.5mm}
$ \mathbf{x}_{switch} \gets$ \FindXSwitchingState{$p_{x,1},~\mathbf{x}_1,~p_{x,2},~\dot{x}_{apex,2}$} \\
\tcp*[f]{Eq.\eqref{eq:vel_x}, \eqref{eq:x_switch}} \\
$ t_{switch} \gets $ \GetTimeAtState{$p_{x,1},~\mathbf{x}_1,~\mathbf{x}_{swtich}$} \tcp*[f]{Eq.\eqref{eq:t_eqn}} \\
$ t_{apex} \gets$ \GetTimeAtState{$p_{x,2},~\mathbf{x}_{apex,2},~\mathbf{x}_{switch}$}\\ \tcp*[f]{Eq.\eqref{eq:t_eqn}} \\
$\mathbf{y}_{switch} \gets$ \Integration{$\mathbf{y}_1, t_{switch}$}  \tcp*[f]{Eq.\eqref{eq:x_state}} \\
$\mathbf{y}_{apex,2} \gets$ \Integration{$\mathbf{y}_{apex,2}, t_{switch}$}  \tcp*[f]{Eq.\eqref{eq:x_state}} \\
$p_{y,2}$ $\gets$     \FindPy{$\mathbf{y}_{switch}, \dot{y}_{apex}, t_{apex}$} \tcp*[f]{Eq.\eqref{eq:yp}}
\end{algorithm}
Let us focus on obtaining the step switching time. We can easily manipulate Eq.~\eqref{eq:x_state} to analytical solve for the time variable,
\begin{equation}
\label{eq:t_eqn}
t  = \frac{1}{\omega} \ln \Big( \frac{x + \frac{1}{\omega}\dot{x} - p_x}{2 A} \Big).
\end{equation}
To find the dynamics, $\dot{x} = f(x)$, which will lead to the switching state solution, let us remove the $t$ term by plugging Eq.~\eqref{eq:t_eqn} into Eq.~\eqref{eq:x_state}.
\begin{equation}
x = A \frac{x + \frac{\dot{x}}{\omega} - p_x}{2 A} + B \frac{2 A}{x + \frac{\dot{x}}{\omega} - p_x } + p_x
\end{equation}
\begin{align}
\frac{1}{2}(x - p_x - \frac{\dot{x}}{\omega}) &=  \frac{2 AB}{x + \frac{\dot{x}}{\omega} - p_x} \\
(x - p_x)^2 - \Big(\frac{\dot{x}}{\omega}\Big)^2  &= 4 AB
\end{align}
By performing some algebra we get, 
\begin{equation}
\label{eq:vel_x}
\dot{x} = \pm \sqrt{\frac{g}{h} \Big( (x - p_x)^2 - (x_{0} - p_x)^2 \Big) + \dot{x}_{0}^2 }.
\end{equation}
Given two phase trajectories associated with consecutive walking steps (e.g. $p_{x,1}$ and $p_{x,2}$), initial states for each (e.g. $\mathbf{x}_{0,1}$ and $\mathbf{x}_{0,2}$) and assuming the robot walks forward (i.e. $\dot{x}_{switch}$) is positive, we calculate the phase space intersection point of each step's CoM trajectory via continuity of velocities from Eq.~\eqref{eq:vel_x}:

\begin{equation}
\label{eq:x_switch}
\begin{split}
x_{\rm switch}&=\frac{1}{2}\Big(  \frac{C}{p_{x,2} - p_{x,1}} + (p_{x,1} + p_{x,2}) \Big)\\
C&=(x_{{0},1}-p_{x,1})^2 - (x_{{0},2} - p_{x,2})^2 + \frac{\dot{x}_{{0},2}^2 - \dot{x}_{{0},1}^2}{\omega^2}
\end{split}
\end{equation}
We can now find the step switching time by plugging the computed switching position into Eqs~\eqref{eq:vel_x} and~\eqref{eq:t_eqn}. In addition, we can obtain the timing at the apex velocity from Eq.~\eqref{eq:t_eqn}. The final step is to find the $y$ directional foot placement. We first calculate $\mathbf{y}_{switch}$ by plugging $t_{switch}$ into the $y$ directional state equation, which has identical form to Eq.~\eqref{eq:x_state}. Then, by using the equality that $\dot{y} (t_{\rm apex})=\dot{y}_{apex}$, we can find $p_{y}$,
\begin{equation}
\label{eq:yp}
\begin{split}
p_y &= \frac{\dot{y}_{apex}-C}{D},\\
C &= \frac{\omega}{2} \big( (y_{switch}+\frac{\dot{y}_{switch}}{\omega})e^{\omega t_{apex}} -  \\
&\quad\quad\quad\quad (y_{switch}-\frac{\dot{y}_{switch}}{\omega})e^{-\omega t_{apex}}\big)\\
D &= \frac{\omega}{2}(e^{-\omega t_{apex}} - e^{\omega t_{apex}})
\end{split}
\end{equation}
After calculating $p_y$, we can easily get $y_{apex}$ and $\dot{y}_{apex}$ by using Eq.~\eqref{eq:x_state}.

\addtolength{\textheight}{-12cm}

\bibliographystyle{IEEEtran}
\bibliography{references}

\end{document}